\documentclass{article}
\usepackage{mathtools}
\usepackage{bbm}
\usepackage{multirow}
\usepackage{booktabs}

\usepackage{times}
\usepackage{graphicx} 
\usepackage{subfigure} 

\usepackage{natbib}

\usepackage{algorithm}
\usepackage{algorithmic}

\usepackage{hyperref}

\usepackage[accepted]{icml2017}

\icmltitlerunning{MultiCheXNet}

\begin{document} 

\twocolumn[
\icmltitle{MultiCheXNet: A Multi-Task Learning Deep Network For Pneumonia-like Diseases Diagnosis From X-ray Scans \footnote{*} }



\icmlsetsymbol{equal}{*}
\icmlsetsymbol{equalequal}{**}

\begin{icmlauthorlist}
\icmlauthor{Abdullah Tarek Farag}{Abdullah}
\icmlauthor{Ahmed Raafat Abd El-Wahab}{Raafat}
\icmlauthor{Mahmoud Nada}{Nada}
\icmlauthor{Mohamed Yasser Abd El-Hakeem}{equal,Yasser}
\icmlauthor{Omar Sayed Mahmoud}{equal,Omar}
\icmlauthor{Reem Khaled Rashwan}{Reem}
\icmlauthor{Ahmad El Sallab}{Sallab}
\end{icmlauthorlist}

\icmlaffiliation{Abdullah}{Speakol, abdullahtarek57@gmail.com}
\icmlaffiliation{Raafat}{Cairo university, A.raafat.pro@gmail.com}
\icmlaffiliation{Nada}{Kafr El Sheikh University, mahmoudnada5997@gmail.com}
\icmlaffiliation{Yasser}{Helwan University, mohamed.y2519@gmail.com}
\icmlaffiliation{Omar}{Helwan University, sayedomar74@gmail.com}
\icmlaffiliation{Reem}{Misr University for Science and Technology, ReemKhaledRashwan@gmail.com}
\icmlaffiliation{Sallab}{Ahmad El Sallab is a Senior Expert of AI, ahmad.elsallab@gmail.com}

\icmlcorrespondingauthor{Ahmad El Sallab}{ahmad.elsallab@gmail.com}

\icmlkeywords{Multi-task Learning, Convolutional Neural Networks, Medical Imaging, Chest X-ray, CXR}

\vskip 0.3in
]



\footnotetext{Code available at \\ https://github.com/ahmadelsallab/MultiCheXNet}
\printAffiliationsAndNotice{} 

\begin{abstract} 
We present MultiCheXNet, an end-to-end Multi-task learning model, that is able to take advantage of different X-rays data sets of Pneumonia-like diseases in one neural architecture, performing three tasks at the same time; diagnosis, segmentation and localization. The common encoder in our architecture can capture useful common features present in the different tasks. The common encoder has another advantage of efficient computations, which speeds up the inference time compared to separate models. The specialized decoders heads can then capture the task-specific features. We employ teacher forcing to address the issue of negative samples that hurt the segmentation and localization performance. Finally, we employ transfer learning to fine tune the classifier on unseen pneumonia-like diseases. The MTL architecture can be trained on joint or disjoint labeled data sets. The training of the architecture follows a carefully designed protocol, that pre-trains different sub-models on specialized datasets, before being integrated in the joint MTL model. Our experimental setup involves variety of data sets, where the baseline performance of the 3 tasks is compared to the MTL architecture performance. Moreover, we evaluate the transfer learning mode to COVID-19 data set, both from individual classifier model, and from MTL architecture classification head.  
\end{abstract} 

\begin{figure*}
\begin{center}
\centerline{\includegraphics[width=170mm]{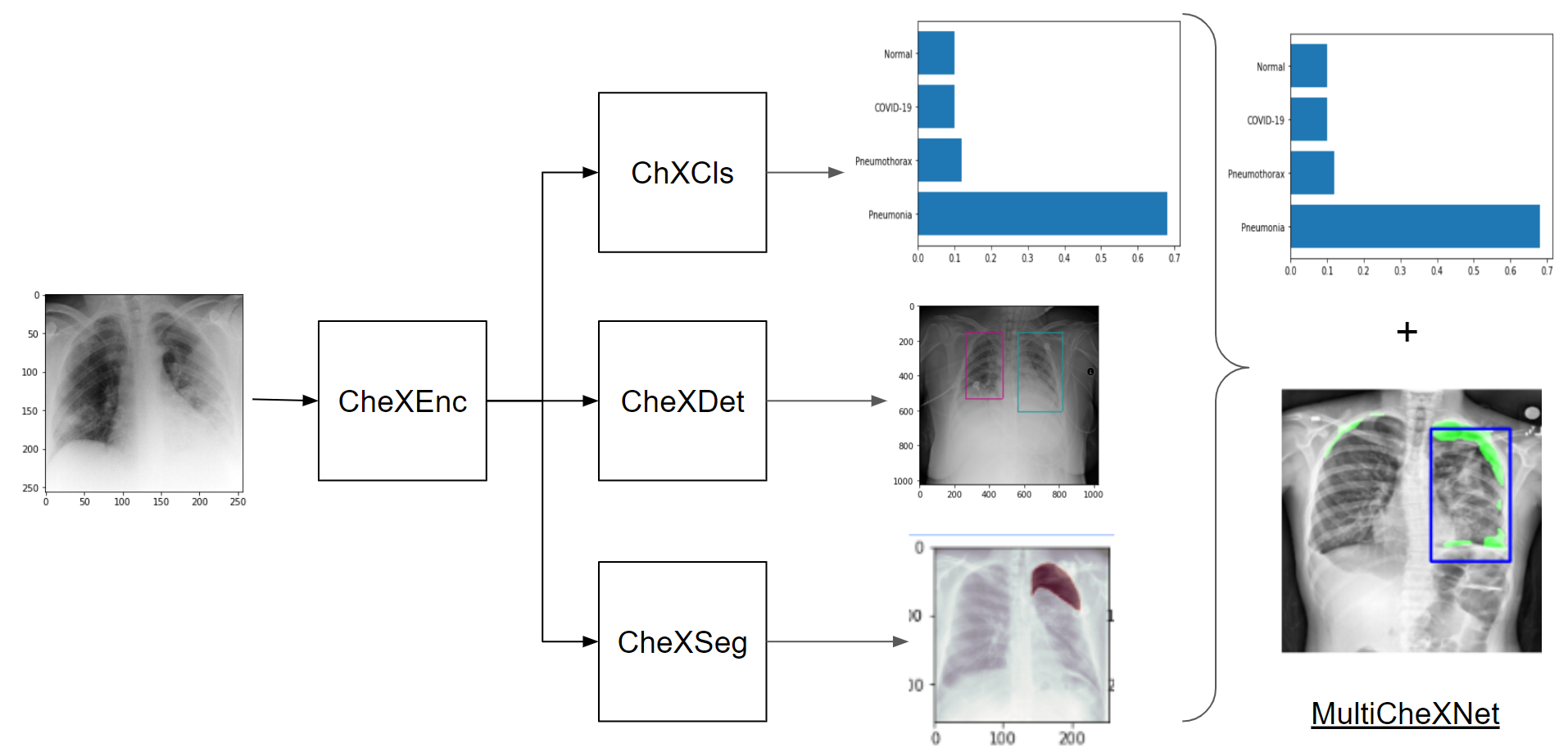}}
\caption{\label{Fig1} MultiCheXNet}
\label{icml-historical}
\end{center}
\end{figure*}

\section{Introduction}
\label{submission}

Diagnosis of Pneumonia-like diseases from X-ray scans has shown promising results \cite{rajpurkar2017chexnet}, leading to Radiologist level performance. The potential of Deep Learning (DL) with Computer Vision (CV) is not only in performance, but also in speed to diagnosed scans. Multi-task learning (MTL) framework has proven to be effective in different domains \cite{teichmann2018multinet}. MTL enables different, yet related, tasks to be performed using one neural architecture model. Not only this paradigm enables efficient use of convolution layers, but also, it improves the performance over the individual tasks models, thanks to the captured common features from many data sets. The challenge is usually in having a jointly labeled data set. This is usually overcome with carefully designed training pipelines, making use of alternating training.

Different diseases can be diagnosed at one from the same X-ray image, thanks to multi-label classification \cite{rajpurkar2017chexnet}. This is somehow a form of multi-tasks learning. While diagnosis of diseases from X-rays is more about classification, other tasks, like segmentation \cite{SIIM-ACR}, and localisation \cite{RSNA}, can be important guidance to radiologists both for interpretation and better diagnosis of the disease. Also, Deep learning is often said to be black box, so model interpret-ability is an important factor in the adoption of these models. Providing segmentation and localization of the disease segment, in addition to the disease class, provides better evidence and enhances interpret-ability.

We propose MultiCheXNet, an MTL architecture comprised of a common convolutional encoder, and three tasks heads; 1) classification, 2) localisation and 3) segmentation. The MTL common encoder and the classifier head, are pre-trained on generic data set \cite{wang2017hospital}, to be able to classify 14 diseases. Also, each of the localisation and segmentation heads are pre-trained before being integrated in the MTL architecture. The final MTL architecture comprising the three heads and the encoder is trained end-to-end. 

Since we do not have joint labels for the three tasks, we perform alternating training, using two data sets, with two heads trained at a time. In another setup, we create localization boxes from the segmentation masks, and perform joint training for the three tasks at once. Moreover, the available datasets for segmentation and detection, are actually for different diseases classes. However, we rely on the fact that X-ray images are somewhat similar and have a standard to some extent. So even if the datasets are for different diseases, the features are likely to be similar, since patterns of diseases are similar across different diseases. This justifies the common encoder importance in the MTL architecture. 

Medical data sets usually suffer class imbalance, due to the nature of the problem having more negative samples than the positive ones. For tasks like segmentation or object detection, this can be hurtful since those models are not good at providing "no detections". We propose a multi-stage pipeline, integrating a first stage classifier that filters negative classes, before performing other tasks. During training, we employ teacher forcing, where we feed ground truth positive classes with certain percent, and classified samples from the first stage classifier for the remaining percent of training samples. 

Finally, we employ transfer learning to unseen Pneumonia-like diseases, from the MTL architecture. Using the common convolutional encoder and the classification head, we transfer the model weights, and fine tune the top layers to classify new unseen, Pneumonia-like diseases. Out experimental use-case is COVID-19 classification. Plugging X-rays from unseen, new disease like COVID-19, into  the MTL architecture, can provide free segmentation and localization boxes of disease findings, without being trained on that particular disease data. This provides a form of few-shot learning, that can be helpful for new diseases where annotated data for segmentation or localisation is still not yet available. This setup makes sense, since the features and symptoms of Pneumonia-like disease are very similar in the X-ray scans, which gives more advantage to the learnt features in the MTL encoder trained from other diseases data sets.

To evaluate MultiCheXNet, we first develop some baselines, for each of the three individual heads. The classification head is trained on \cite{wang2017hospital}, the segmentation on \cite{SIIM-ACR} and the localisation on \cite{RSNA}. In the localisation and segmentation models, teacher forcing is performed, with the ground truth positive cases, and with some percent from the pre-trained classifier output. Moreover, each head classifier is further pre-trained on its own data set. After integrating and training the MTL model, we compare each head performance, with their individual performance, on the  same data sets they were pre-trained on, to show the effect of MTL. Finally, we develop a baseline for an individual classifier on \cite{cohen2020covid}, and compare that to the performance of the MTL head on the same COVID-19 data set.

\section{Related Work}
\label{submission}

\textbf{CheXNet} \cite{rajpurkar2017chexnet} was one of the pioneering works that applied Deep Learning and Convolutional networks to medical imaging diagnosis \cite{lakhani2017deep}, \cite{wang2017hospital}. The authors claim radiologist level accuracy. This inspired many following works to apply deep learning in medical images diagnosis, extending to tasks like semantic segmentation and object detection and localization \cite{sirazitdinov2019deep}, \cite{jaiswal2019identifying}. 

Recently, many efforts have been made to apply the same idea to diagnose COVID-19 disease from X-rays and CT scans \cite{wang2020covid},  \cite{basu2020deep}, \cite{maguolo2020critic}, \cite{tartaglione2020unveiling}. These efforts were usually constrained by the limited amount of data, due to the recency of the disease outbreak, \cite{maguolo2020critic}, \cite{tartaglione2020unveiling}.

In the context of computer vision in automated driving, \textbf{MultiNet} \cite{teichmann2018multinet} is also a pioneering work applying Multi-task learning for real-time object detection and road segmentation, making use of the idea of shared convolutional encoder to reduce unnecessarily redundant computation if separate models were developed for each task. Further, it was shown that overall detection and segmentation performance improved over the individual models trainings.

\section{Approach}
\label{submission}

\subsection{Multi-Task Learning architecture}
\begin{figure*}
\begin{center}
\centerline{\includegraphics[width=120mm]{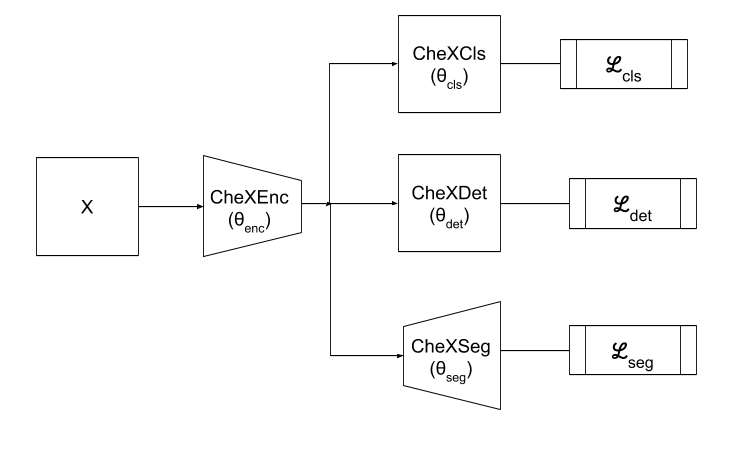}}
\caption{\label{Fig2} MTL architecture}
\label{icml-historical}
\end{center}
\end{figure*}

MultiChexNet is a Multi-Task Learning (MTL) architecture, of four main neural network blocks; a common encoder, and three decoders, as shown in \ref{Fig2}. We collectively define each block as a function $F$ parametrized with some weights $\theta$; $F(X;\theta)=F_{\theta}(X)$, where $X \in \mathbbm{R}^{M \times L \times W}$ is the input data of $M$ grey scaled X-ray images, with each image $x \sim X$ of size $L \times W$.
\subsection{Encoder (CheXEnc)}

The encoder network CheXEnc is a convolutional neural network, with a set of trainable kernel weights;  $\theta_{enc}$, that encodes the input X-ray scan $X$ into a common representation vector $X_{enc} \in \mathbbm{R}^{M \times N \times d \times d}$, where $N$ is the number of feature maps, and $d$ is the dimension of the encoded feature map:

\begin{eqnarray}
\label{encoder}
    X_{enc} = F_{\theta_{enc}}(X)
\end{eqnarray}

Three tasks are to be learned using three specialized convolutional decoders: CheXCls, CheXDet and CheXSeg. Each head is associated to as loss function for each task: $\mathcal{L}_{cls}$, $\mathcal{L}_{det}$ and $\mathcal{L}_{seg}$. The whole architecture is optimized end-to-end minimizing the total loss. The common encoder will be updated as part of the three losses, and hence will capture the common features, and will enable to transfer the learned features between the three tasks, which improves the overall performance. Another advantage is that, a single inference pass through the encoder is required, instead of three separate decoders in case of individual networks, which save the inference time and reduces the final model size.

Following are the details of each decoder, the associated loss, and the overall loss.

\subsection{Classification decoder head (CheXCls)}
The classification head decoder is a convolutional neural network parametrized by $\theta_{cls}$ kernel weights, that operates on the encoded feature map from CheXEnc:

\begin{eqnarray}
\label{Y_cls}
\hat Y_{cls} = F_{\theta_{cls}}(X_{enc})
\end{eqnarray}

Where $\hat Y_{cls} \in \mathbbm{R}^{M \times C}$ represent $M$ output vectors for the input $X$ X-ray images; such that each output vector $\hat y \sim \hat Y_{cls} ; \hat y \in \mathbbm{R}^{1 \times C}$ is the result of $C$ output neurons with sigmoid activations, each associated to a class label.
We collectively refer to the dataset used for classification as $D_{cls} = (X, Y_{cls})$, where $Y_{cls} \in \mathbbm{R}^{M \times C}$ are the ground truth labels, and $y \sim Y_{cls} ; y \in \mathbbm{R}^{1 \times C}$ are the individual samples labels. The loss of classification is a binary cross entropy loss over each output class neuron as follows:

\begin{eqnarray}
\label{L_cls}
\begin{multlined}
\mathcal{L}_{cls}(D_{cls}; \theta_{enc}, \theta_{cls}) = \\
\mathbbm{E}_{x,y \sim D_{cls}}[\sum_{i=1}^{C} y_{i} \times log(\hat{y}_i) + (1-y_{i}) \times log(1- \hat{y}_i) ]
\end{multlined}
\end{eqnarray}

\subsection{Detection decoder head (CheXDet)}
The detection head decoder is a convolutional neural network parametrized by $\theta_{det}$ kernel weights, that operates on the encoded feature map from CheXEnc:

\begin{eqnarray}
\label{Y_det}
\hat Y_{det} = F_{\theta_{det}}(X_{enc})
\end{eqnarray}

Where $\hat Y_{det} \in \mathbbm{R}^{M \times K \times 4}$ represent $M$ outputs corresponding the input $X$ X-ray images. Each output $\hat y \sim Y_{det} ; \hat y \in \mathbbm{R}^{K \times 4}$ represents output $K$ anchor boxes. Every box parameters are resulting from 4 output neurons with ReLU activations.

We refer to the dataset used for detection as $D_{det} = (X, Y_{det})$, where $Y_{det} \in \mathbbm{R}^{M \times B \times 4}$, and $y \sim Y_{det}; y \in \mathbbm{R}^{B \times 4}$ represent $B$ ground truth bounding boxes for an input image $x \sim X$. $B$ is the maximum number of ground truth bounding boxes in an input image. Each bounding box ($\hat y_i$ or $y_i$) is represented with four numbers: ${c_x, c_y, l, w}$. Where $c_x$ and $c_y$ are the coordinates of the box center, while $l$ and $w$ are the length and width of the box. The loss of detection is a L2 regression over the box parameters of both the output anchors and the ground truth boxes, provided an object exists in the region of overlap between both boxes:

\begin{eqnarray}
\label{L_det}
\begin{multlined}
\mathcal{L}_{det}(D_{det}; \theta_{enc}, \theta_{det}) = \\
\mathbbm{E}_{x,y \sim D_{det}}[\sum_{i=1}^{K} \sum_{j=1}^{B} \mathbbm{1}_{ij}^{obj}|| y_{i} - \hat{y}_{i}||^2 ]
\end{multlined}
\end{eqnarray}

\subsection{Segmentation decoder head (CheXSeg)}

The segmentation head decoder is a convolutional neural network parametrized by $\theta_{seg}$ kernel weights, that operates on the encoded feature map from CheXEnc:

\begin{eqnarray}
\label{Y_seg}
\hat Y_{seg} = F_{\theta_{seg}}(X_{enc})
\end{eqnarray}

Where $\hat Y_{seg} \in \mathbbm{R}^{M \times L \times W}$ is the set of $M$ output segmentation masks for the input $X$ X-ray images. Each output mask $\hat y \sim Y_{det} ; \hat y \in \mathbbm{R}^{L \times W}$ represents a binary mask over the image pixels (result of sigmoid activation), segmenting the area of the disease.

We collectively refer to the dataset used for classification as $D_{seg} = (X, Y_{seg})$, where $Y_seg \in \mathbbm{R}^{M \times L \times W}$ represent $M$ output masks, each $y \sim Y_{seg}; y \in \mathbbm{R}^{L \times W}$ is a binary mask over the input image $x \sim X$. The loss of detection is a binary cross entropy over each value of the mask outputs:

\begin{eqnarray}
\label{L_seg}
\begin{multlined}
\mathcal{L}_{seg}(D_{seg}; \theta_{enc}, \theta_{seg}) = \\
\mathbbm{E}_{x,y \sim D_{seg}}[\sum_{i=1}^L \sum_{j=1}^W  y_{ij} \times log(\hat{y}_{ij})             \\ + (1-y_{ij}) \times log(1- \hat{y}_{ij})]
\end{multlined}
\end{eqnarray}

\subsection{Overall loss}
Given a jointly labelled dataset: $D=(X,Y)$, where $Y= \{ Y_{cls}, Y_{det}, Y_{seg} \}$ is the set of joint labels for the class, bounding boxes and segmentation masks annotations for the same input $X$, we can calculate the total loss as:
\begin{eqnarray}
\label{joint_loss}
\begin{multlined}
\mathcal{L}_{tot}(D;\theta) = \\
\mathcal{L}_{cls}(D_{cls};\theta_{enc}, \theta_{cls}) + \\ \mathcal{L}_{det}(D_{det};\theta_{enc}, \theta_{det}) + \\ \mathcal{L}_{seg}(D_{seg};\theta_{enc}, \theta_{seg})
\end{multlined}
\end{eqnarray}

Where $\theta = \{\theta_{enc}, \theta_{cls}, \theta_{det}, \theta_{seg} \}$ can be optimized end-to-end minimizing the total loss as follows:

\begin{eqnarray}
\label{joint_params}
\begin{multlined}
\theta^{*} = argmin_{\theta} \mathcal{L}_{tot}
\end{multlined}
\end{eqnarray}

However, having a joint dataset $D$ is not an always easy. In some cases, if we have segmentation masks, we can have the corresponding boxes. The opposite will result in wide masks, spanning the box area. We will discuss other alternative training method in the training protocol section.

\subsection{Positive cases classifier}

We have noticed during training of the segmentation and detection heads that, feeding both positive and negative samples hurts the performance. This is mostly because of the high data imbalance, where most of the input X-ray scans correspond to negative cases, where no bounding box or segmentation mask available. This creates some difficulty for these decoders, specially the detection head, where it is by default generating boxes, and then filtered by post processing.

To avoid this issue, we train the segmentation and detection heads on positive samples only (the ones with mask or box in the label). Now the question is: during inference, how we control feeding only positive samples?. 

For that we design a first stage classifier $Cls$, trained as a positive/negative classifier, followed by the MTL encoder and decoders. The classifier will filter the negative cases, and prevent them from going through the segmentation or detection heads. The classifier can be trained independently as a binary classifier with binary cross entropy loss. During MTL training, this classifier is frozen (appears as grey box in the architecture diagrams).

However, we can use the same classification head that we train as part of the MTL architecture, as a filter in that pipeline. In that case the input $X$ is first passed on the CheXEnc+CheXCls sub-network, so that the negative cases are filtered out if no disease from the set of $C$ diseases is detected by CheXCls decoder. Otherwise, it is a positive case, and can be passed to the segmentation and detection heads.

The drawback of this approach is that we need to run the encoder twice; once for filtering, and then through the MTL. This can be optimized by caching the encoder output, and using it for both phases of the pipeline.

\subsection{Teacher forcing(TF)}

Having only positive cases during training of the MTL decoders is a form of teacher forcing (TF). During training, the first stage classifier $Cls$, needs to pre-trained independently on negative cases, before being trained only on positive cases as part of MTL architecture. This will be detailed in the experimental setup section with the training phases.

Another drawback could be that, the detection and segmentation heads are never trained on negative cases. This is an issue if the first stage classifier fails and passes a negative case by mistake. For that, we mix in training to have some negative cases for some percent of the training data, and also some samples as classified by the first stage classifier. This is handled by tossing a coin with every batch or sample, with probability $p$ forcing the positive sample from the ground truth, and $1-p$ to take whatever passed from the first stage classifier $Cls$. In the architecture diagrams, we set the $Cls$ module as "dotted" as a notation of the coin tossing in teacher forcing training.

\subsection{Transfer Learning to unseen diseases}
The classification task is the mostly needed for diagnosis, while segmentation and detection are more needed for interpret-ability and explanability of the diagnosis. There are many data sets with one or more disease. Such datasets are collected over long time. For new diseases, limited amount of data can be available, specially during the initial outbreak, like in the case of COVID-19. Since the classification head has a multi-label loss, so transfer learning can be employed to fine-tune the separate diseases labels. Under MTL architecture, we can further make use of the common features captured in the CheXEnc network. The classification path (CheXEnc+CheXCls) can be further fine tuned on new data sets with new diseases.

\section{Experimental Setup} 
\subsection{Training protocol}
As explained, the MultiCheXNet architecture can be trained end-to-end, given a jointly labeled dataset for all the three tasks. However, there are some aspects we consider:

\begin{itemize}
    \item Limited jointly labeled datasets: as explained before, it is possible to obtain bounding boxes from segmentation masks, but the opposite is inefficient. We have many data sets available, some with classification only, some with segmentation masks annotations and others with boxes annotations. We want to make efficient use of all of them.
    
    \item Big-bang integration: we have four models, with four sets of weights. Instead of initializing them with  random weights, we choose to pre-train each head alone, and then integrate and fine tune them in the MTL architecture. This will help all the models to start from the baseline performance of individual networks, and then improve thanks to the common features captures in the common encoder. The common encoder is common in all pre-training phases.
    
    \item The first stage classifier $Cls$ needs to be trained alone first, before being used  for teacher forcing training. Where in MTL training of either segmentation or detection heads, teacher forcing will mostly pass the positive cases, so the $Cls$ will not be enough trained on negative cases, while this is needed for its role as a filter. Hence, we need to pre-train this classifier alone at different stages. We will pre-train it on a generic dataset \cite{wang2017hospital}, and also we pre-train it within the detection or segmentation pre-training, on all the positive and negative classes, even before the detection and segmentation heads are trained on positive cases. 
    
\end{itemize}

The phases are as shown in figure \ref{Protocol}:
\begin{itemize}
    \item Classifier pre-training: CheXEnc ($\theta_{enc}$) + CheXCls ($\theta_{cls}$).
    \item Detection head pre-training: CheXEnc ($\theta_{enc}$) + CheXDet ($\theta_{det}$).
    \item Segmentation head pre-training CheXEnc ($\theta_{enc}$) + CheXSeg ($\theta_{seg}$). 
    \item MTL training CheXEnc ($\theta_{enc}$) + CheXCls ($\theta_{cls}$) + CheXSeg ($\theta_{seg}$) + CheXDet ($\theta_{det}$).     
    \item Transfer learning to new data/diseases: CheXEnc ($\theta_{enc}$) + CheXCls ($\theta_{cls}$).
    
\end{itemize}

The first 3 phases are considered as a baseline for comparison of individual performance against the integrated MTL architecture. For all the below figures, Grey boxes means not-trainable, dotted boxes means teacher forcing with percentage.

\subsubsection{Classification head pre-training}
The classification sub-network: CheXEnc ($\theta_{enc}$) + CheXCls ($\theta_{cls}$) is pre-trained on ChestX-ray-14 (ref) for 14 classes. This sub-network will further serve for the pipeline classifier to filter out the negative cases.

\begin{eqnarray}
\label{Cls_pretrain}
\begin{multlined}
\theta_{enc}, \theta_{cls} = argmin_{\theta_{enc}, \theta_{cls}} \mathcal{L}_{cls}(D_{cls};\theta_{enc}, \theta_{cls}) 
\end{multlined}
\end{eqnarray}

\subsubsection{Detection head pre-training}
In the detection head pre-training,the following networks are updated on RSNA dataset (ref): CheXEnc ($\theta_{enc}$) + CheXDet ($\theta_{det}$). The encoder is taken from the pre-trained classifier in the first phase. During this phase, we need to run the pipeline classifier to filter out the negative cases. For that, we further pre-train the classifier on the specific positive/negative cases from RSNA dataset.

Also, we employ teacher forcing with $p=90\%$, which means that we toss a coin with every batch, and 10\% of the time we take the output of the pipeline classifier, and 90\% of the time we force the positive class samples.

The detection dataset \cite{RSNA}, includes both the boxes, in addition to the disease classification of the overall image (e.g. Pneumonia). Before pre-training the detection head, we first pre-train the classification path on the classification targets of the dataset. The reason is that; we will only focus on the positive cases during the detection pre-training, and hence the classifier will not be enough trained on negative cases. For that we create a new dataset out of $D_{det}$; $D_{det}^{cls}=(X, Y_{det}^{cls})$, where $Y_{det}^{cls}$ are the classes of the images of the detection datast $D_{det}$:

\begin{eqnarray}
\label{L_cls}
\begin{multlined}
\theta_{enc}, \theta_{cls} = argmin_{\theta_{enc}, \theta_{cls}} \mathcal{L}_{cls}(D_{det}^{cls};\theta_{enc}, \theta_{cls})
\end{multlined}
\end{eqnarray}

Following the pre-training, the encoder and the detection head are further trained on the detection labels $D_{det}$:

\begin{eqnarray}
\label{L_cls}
\begin{multlined}
\theta_{enc}, \theta_{det} = argmin_{\theta_{enc}, \theta_{det}} \mathcal{L}_{det}(D_{det};\theta_{enc}, \theta_{det}) 
\end{multlined}
\end{eqnarray}

\subsubsection{Segmentation head pre-training}
In the segmentation head pre-training,the following networks are updated on SIIM-ACR dataset (ref): CheXEnc ($\theta_{enc}$) + CheXSeg ($\theta_{seg}$). The encoder is taken from the pre-trained classifier in the first phase. During this phase, we need to run the pipeline classifier to filter out the negative cases. For that, we further pre-train the classifier on the specific positive/negative cases from SIIM-ACR dataset.

Teacher forcing is also employed, same as in detection head pre-training. Also, similar to the detection case, we first pre-train the classifier on a new dataset out of $D_{seg}$; $D_{seg}^{cls}=(X, Y_{seg}^{cls})$, where $Y_{seg}^{cls}$ are the classes of the images of the detection datast $D_{seg}$:

\begin{eqnarray}
\label{L_cls}
\begin{multlined}
\theta_{enc}, \theta_{cls} = argmin_{\theta_{enc}, \theta_{cls}} \mathcal{L}_{cls}(D_{seg}^{cls};\theta_{enc}, \theta_{cls}) 
\end{multlined}
\end{eqnarray}

Following the pre-training, the encoder and the segmentation head are further trained on the detection labels $D_{seg}$:

\begin{eqnarray}
\label{L_cls}
\begin{multlined}
\theta_{enc}, \theta_{seg} = argmin_{\theta_{enc}, \theta_{seg}} \mathcal{L}_{seg}(D_{seg};\theta_{enc}, \theta_{seg}) 
\end{multlined}
\end{eqnarray}

\subsection{MTL training}
\begin{figure}
\begin{center}
\centerline{\includegraphics[width=90mm]{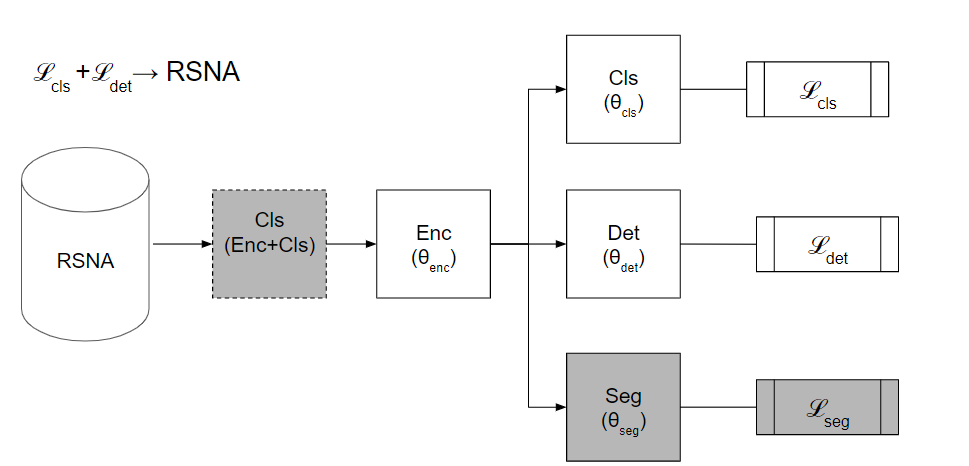}}
\centerline{\includegraphics[width=90mm]{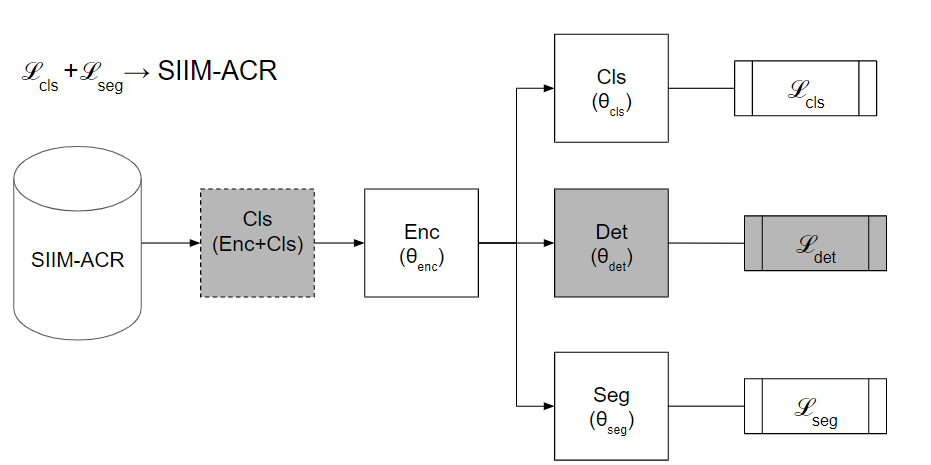}}
\caption{\label{Alt-MTL} Alternating MTL}
\label{icml-historical}
\end{center}
\end{figure}

\begin{figure*}
\begin{center}
\centerline{\includegraphics[width=150mm]{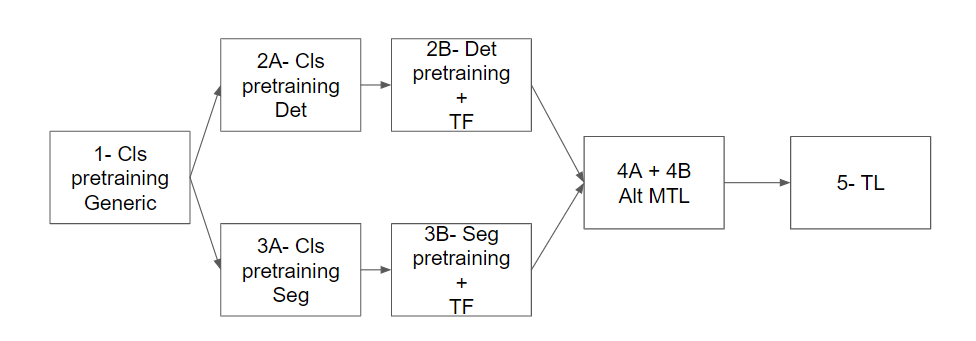}}
\caption{\label{Protocol} Training protocol}
\label{icml-historical}
\end{center}
\end{figure*}

In MTL training, all the four networks are updated: CheXEnc ($\theta_{enc}$) + CheXCls ($\theta_{cls}$) + CheXSeg ($\theta_{seg}$) + CheXDet ($\theta_{det}$). We have two experimental setups:

\begin{itemize}
    \item Joint training: we do not have one training dataset that includes the segmentation masks and bounding boxes, for the same input. However, we could fit a box on the segmentation masks of the SIIM-ACR dataset, and train the model jointly end-to-end. However, this limits the training to only one dataset as in equation \ref{joint_loss}.
    
    \item Alternating training: to handle the lack of jointly labeled datasets for segmentation and detection, we train the segmentation pipeline on SIIM-ACR dataset, alternating with the detection pipeline on RSNA dataset, one batch from each data set at a time. While training one head, the other is not trained as in figure \ref{Alt-MTL}. The common encoder is trained in both steps, and hence improved from both datasets, and can capture the common features. The full algorithm is detailed in algorithm table \ref{alg_alt_mtl}.

\end{itemize}

\begin{algorithm}
    \caption{Alternating MTL}
    \label{alg_alt_mtl}
  \begin{algorithmic}[1]
    \INPUT Datasets $D_{cls}$, $D_{det}$ and $D_{seg}$
    \OUTPUT Optimized model parameters. $\theta=\{\theta_{enc},\theta_{cls},\theta_{det}, \theta_{seg}\}$ 
    \STATE \textbf{Initialization} ‎$\theta^{(0)} = \mathcal{N}(0, I)$
    
    \STATE \textbf{Pre-train $Cls$} ‎$\theta_{enc}, \theta_{cls} = argmin_{\theta_{enc}, \theta_{cls}} \mathcal{L}_{cls}(D_{cls};\theta_{enc}, \theta_{cls})$
    
    \STATE \textbf{Pre-train detection classifier} ‎$\theta_{enc}, \theta_{cls} = argmin_{\theta_{enc}, \theta_{cls}} \mathcal{L}_{cls}(D_{det}^{cls};\theta_{enc}, \theta_{cls})$ 
    
    \STATE \textbf{Pre-train detection head}‎$\theta_{enc}, \theta_{det} = argmin_{\theta_{enc}, \theta_{det}} \mathcal{L}_{det}(D_{det};\theta_{enc}, \theta_{det})$  
    
    \STATE \textbf{Pre-train segmentation classifier} ‎$\theta_{enc}, \theta_{cls} = argmin_{\theta_{enc}, \theta_{cls}} \mathcal{L}_{cls}(D_{seg}^{cls};\theta_{enc}, \theta_{cls})$ 
    
    \STATE \textbf{Pre-train segmentation head}‎$\theta_{enc}, \theta_{seg} = argmin_{\theta_{enc}, \theta_{seg}} \mathcal{L}_{seg}(D_{seg};\theta_{enc}, \theta_{seg})$ 
    
    \WHILE{$i \leq n_{steps}$}
    
    \STATE Sample $D_{batch} \sim D_{det}$
    
    \STATE $\theta_{enc}, \theta_{det} = argmin_{\theta_{enc}, \theta_{det}} \mathcal{L}_{det}(D_{batch};\theta_{enc}, \theta_{det})$ 
    
    \STATE Sample $D_{batch} \sim D_{det}$
    
    \STATE $\theta_{enc}, \theta_{seg} = argmin_{\theta_{enc}, \theta_{seg}} \mathcal{L}_{seg}(D_{batch};\theta_{enc}, \theta_{seg})$
    
    \ENDWHILE
  \end{algorithmic}
\end{algorithm}

\subsection{Transfer learning/ Few-shot learning}
We experiment on the COVID-19 use case, were limited data is available for COVID X-rays. However, we can transfer the pre-trained (CheXEnc+CheXCls) network, on the three heads, on abundant data for different diseases. After the MTL architecture is trained, the classification sub-network CheXEnc ($\theta_{enc}$) + CheXCls ($\theta_{cls}$) is fine-tuned on COVID-19 dataset \cite{cohen2020covid}. The COVID-19 dataset is small, and hence we compare the performance before and after transfer learning. Also, we compare the performance of transfer from the pre-trained classifier in the first stage (only the classification sub-network on ChestX-14 dataset), to COVID-19 data, versus transfer from the MTL architecture.

We fine-tune the classification path $Cls$. For that, we factorize its parameters as follows:

\begin{eqnarray}
\theta_{cls} = [\theta_{feats} ; \theta_{dec}]
\end{eqnarray}

Where $\theta_{feats}$ are the early layers leading to the extracted features, of dimensions $d$ that will be used for classification, and $\theta_{dec} \in \mathbbm{R}^{d \times C}$ are the final set of weights, producing the final output for $C$ classes, resulting from $C$ sigmoid activation neurons. $\theta_{feats}$ are transfered from the pre-trained $\theta_{cls}$ in the MTL algorithm, while $\theta_{dec}$ is fine-tuned from scratch on the new class(es), e.g. COVID-19.

\subsection{Data sets}
\textbf{Chest X-ray14} dataset by \cite{wang2017hospital} has 112,120 annotated for 14 different diseases labels.

\textbf{RSNA} \cite{RSNA} is a Kaggle competetion for detecting and localizing Pneumonia in X-ray lung scans. the data consists of 26,700 scans for training and another 3000 scans for testing, labelled with bounding boxes coordinates of the locations of Pneumonia in the images.

\textbf{SIIM-ACR} \cite{SIIM-ACR}is a Kaggle competetion for detecting and localizing Pneumonia in X-ray lung scans. the data consists of 12,100 scans for training and another 3205 scans for testing, labelled with segmentation masks of the locations of Pneumonia in the images.

\textbf{COVID-19} \cite{cohen2020covid} is a  public open dataset of chest X-ray and CT images of patients which are positive or suspected of COVID-19 and other 16 diseases like MERS, SARS, and ARDS. This dataset is very small (~100 images).

\section{Results}

\begin{figure*}
\begin{center}
\centerline{\includegraphics[width=160mm]{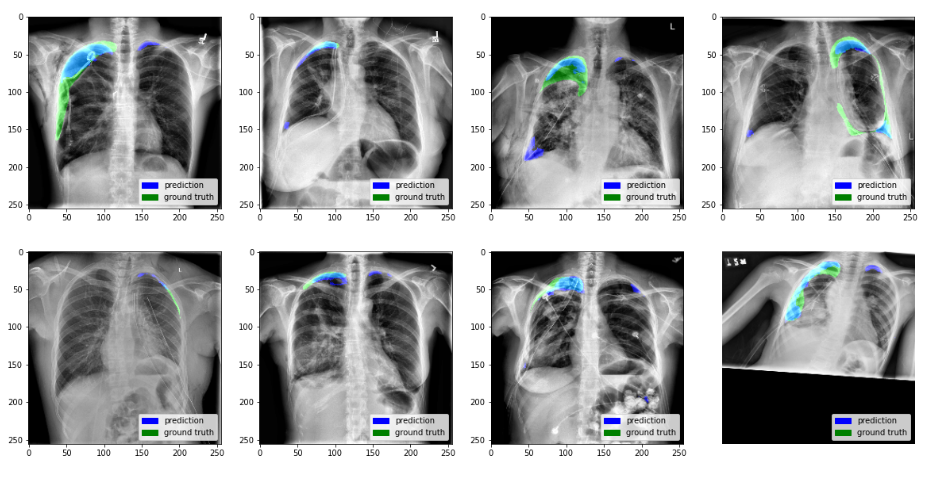}}
\caption{\label{Masks_pred} Samples of segmentation (CheXSeg) output}
\label{icml-historical-1}
\end{center}
\end{figure*}

\begin{figure*}
\begin{center}
\centerline{\includegraphics[width=180mm]{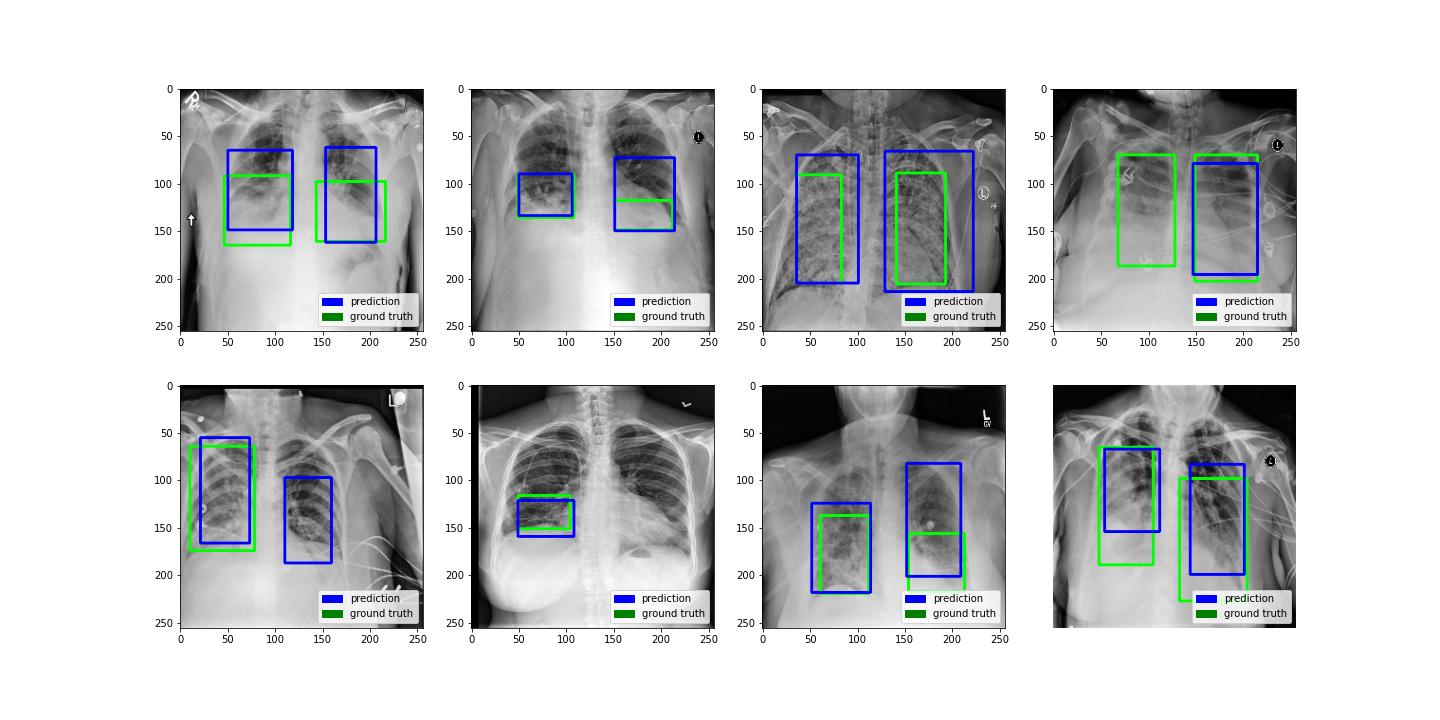}}
\caption{\label{Box_pred} Samples of (detection) CheXDet output}
\label{icml-historical-2}
\end{center}
\end{figure*}

\textbf{Effect of Multi-task learning against separately trained tasks} - In this experiment, we aim at testing the effect of MTL training over the individually trained models on the the three separate tasks. MTL beats all baselines that are individually trained as shown in Table \ref{tab:MTL-res}.

\begin{table*}
  \begin{center}
    \caption{MTL Results with comparison to baselines of individually trained heads (95\% CI)}
    \label{tab:MTL-res}
    \begin{tabular}{c|c|c|c|c|c} 
      \textbf{Experiment} & \textbf{Dataset} & \multicolumn{2}{c|}{\textbf{Classification}} & \textbf{Segmentation} & \textbf{Detection}\\\cline{3-6}

                                    &                               & \textbf{Acc} & \textbf{F1} & \textbf{DICE} & \textbf{mAP}\\
       \hline
       
       Baseline Classification & Chest X-ray 14 & \textbf{0.87 (0.86, 0.88)} & 0.31 (0.27, 0.35)  & - & - \\
       Baseline Segmentation & SIIM-ACR & - & - & 0.68 (0.66, 0.70) & - \\
       Baseline Detection & RSNA & - & - & - & 0.16 (0.15, 0.17) \\
       \hline
       MTL w/o pre-training  & RSNA + SIIM-ACR & 0.52 (0.44, 0.59) & 0.53 (0.43, 0.61) & 0.29 (0.28, 0.30) & \textbf{0.21 (0.20, 0.22)} \\
       (BigBang) & & & & & \\
       \hline
       MTL w/ pre-training  & RSNA + SIIM-ACR & \textbf{0.73 (0.66, 0.79)} & \textbf{0.73 (0.65, 0.80)} & \textbf{0.75 (0.74, 0.76)}  & 0.16 (0.15, 0.17) \\ 
       (\textbf{MultiCheXNet}) & & & & & \\
       \hline

    \end{tabular}
  \end{center}
\end{table*}
\textbf{Effect of pre-training} - The effect of MTL+pre-training, which we tag as MultiCheXNet in Table \ref{tab:MTL-res}, is significant over MTL from scratch, using random initialization, which we call BigBang MTL in Table 1. Since the individual heads are pre-trained on relevant datasets, they tend to perform better when integrated in the MTL model, and helps reducing the tuning effort of the overall model, since

\textbf{Effect of Teacher Forcing and Positive Cases classifier} - With the inclusion of the positive cases classifier, which acts as a filter, some errors might occur, and some negative cases can pass to the segmentation and detection heads by mistake as false positives (those should have been filtered by the first stage classifier). Table \ref{tab:MTL-TF} shows the effect of those false positives on the overall results. It is clear that the results drop on all metrics of the 3 heads, however, given the already good performance of the first stage classifier, the effect is not critical.

\begin{table*}
  \begin{center}
    \caption{Effect of Teacher Forcing and Positive Cases classifier (95\% CI)}
    \label{tab:MTL-TF}
    \begin{tabular}{c|c|c|c|c|c} 
      \textbf{Experiment} & \textbf{Dataset} & \multicolumn{2}{c|}{\textbf{Classification}} & \textbf{Segmentation} & \textbf{Detection}\\\cline{3-6}

                                    &                               & \textbf{Accuracy} & \textbf{F1} & \textbf{DICE} & \textbf{mAP}\\
       \hline
       MTL w/ & RSNA + SIIM-ACR & 0.77 (0.70, 0.82) & 0.77 (0.70, 0.83) & 0.78 (0.77, 0.79) & \textbf{0.19 (0.18, 0.20)} \\
       100\% Teacher Forcing & & & & & \\
       \hline
       MTL w/ pre-training  & RSNA + SIIM-ACR & 0.73 (0.66, 0.79) & 0.73 (0.65, 0.80) & 0.75 (0.74, 0.76) & 0.16 (0.15, 0.17) \\
       (\textbf{MultiCheXNet}) & & & & & \\
       \hline

    \end{tabular}
  \end{center}
\end{table*}

\textbf{Multi-task learning – Transfer learning (MTL-TL) scenario} - The effect of transferring the learnt classifier (CheXCls), and fine tuning to new unseen classes is demonstrated in Table \ref{tab:MLT-TL-COVID}, on COVID-19 dataset, which has only around 100 images, and suffer from high class imbalance towards the negative class as expected. After fine-tuning on the new dataset, the pre-trained classifier performs significantly better than the baseline, trained from scratch on the small COVID-19 dataset alone.
\begin{table*}
  \begin{center}
    \caption{Transfer learning from MTL classifier head and fine tuning to new diseases (COVID-19) (95\% CI)}
    \label{tab:MLT-TL-COVID}
    \begin{tabular}{c|c|c|c|c} 
       & \textbf{Source Dataset} & \textbf{Target Dataset} & \textbf{Accuracy} & \textbf{F1}\\
       \hline
       Baseline COVID & - & COVID & 0.59 (0.52, 0.66) & 0.72 (0.63, 0.76) \\
       
       MTL - TL: MultiCheXNet (CheXCls) & RSNA + SIIM-ACR & COVID & \textbf{0.70 (0.64, 0.76)} & \textbf{0.80 (0.75, 0.85)} \\

       \hline

    \end{tabular}
  \end{center}
\end{table*}

\textbf{Comparison to CheXNet \cite{rajpurkar2017chexnet}} - We treat CheXNet as a benchmark to compare our results to, since it is the closest work to ours. To do that, we take the trained classification head (CheXCls) in MultiCheXNet, and test its results on the Chest X-ray 14 dataset \cite{wang2017hospital} in order to compare our results to CheXNet. Table \ref{tab:CheXNet} shows the improvement of F1 score of  MultiCheXNet (CheXCls) over the CheXNet. In our setup, we treat CheXNet architecture as a multi-label architecture, with binary cross entropy loss and sigmoid neurons on all the 14 classes of Chest X-ray 14 dataset \cite{wang2017hospital}. The reason is that, we treat our classification head as a "universal" classifier, that can be fine tuned jointly on any dataset, and produce joint disease classifications for all classes at once. This is different from the original CheXNet \cite{rajpurkar2017chexnet} which is based on binary classification setup, handling one disease at a time, focused on Pneumonia, and tested in the same fashion on other diseases. Thus, CheXNet \cite{rajpurkar2017chexnet} uses one-versus-all setup, with binary cross entropy loss and only one sigmoid, with positive case (1) for the classified disease and negative (0) for "all" others. So we train our own version of CheXNet, called CheXNet-Multilabel in Table \ref{tab:CheXNet}, with multi-label setup, and compare that to the MultiCheXNet architecture with the same multi-label problem setup on Chest X-ray 14 dataset \cite{wang2017hospital} for fair comparison.

\begin{table*}
  \begin{center}
    \caption{Comparison to CheXNet (95\% CI)}
    \label{tab:CheXNet}
    \begin{tabular}{c|c|c|c} 
       & \textbf{Source Dataset} & \textbf{Target Dataset} & \textbf{F1} \\
       \hline
       CheXNet-Multilabel & - & Chest X-ray 14 & 0.31 (0.27, 0.35) \\
       
       MultiCheXNet (CheXCls) & RSNA + SIIM-ACR & Chest X-ray 14 & \textbf{0.36 (0.31, 0.39)}\\

       \hline

    \end{tabular}
  \end{center}
\end{table*}

As can be seen in \ref{tab:CheXNet}, fine tuning MultiCheXNet classification pipeline (CheXCls) outperfrom the baseline CheXNet architecture, which again proves the added value of MTL training.

\section{Conclusion and Future work}

In this work, we presented MultiCheXNet, an MTL framework for Pneumonia-like diseases diagnosis in X-ray scans. Our model can perform three diagnosis tasks at a time, using one model; classification, segmentation and detection of the disease area. Moreover, we carefully designed a training protocol to help the convergence of the overall MTL architecture, making use of all the available datasets, although being dis-jointly labeled for the tasks. This opens the door to incorporating more datasets as needed and available. Finally, we demonstrated the ability to transfer the learned models to new diseases classes, which might suffer the scarcity of data, like COVID-19 case.
Future work shall include careful Ablation study of the effect of positive cases classifier, effect of teacher forcing, per disease class evaluation and better statistical evaluation methodology.

This is a work in progress, representing a step towards employing MTL techniques in computer vision aided diagnosis from radiology images. However, it suffers some issues in the evaluation methodology that the authors are aware of, and to be treated in future works. For this reason, this work is not intended to be used as is, or deployed in real treatment protocols, diagnosis or hospitals, due to the lack of enough experimentation and statistically sound evaluation methodologies.

\bibliography{main}

\begin{thebibliography}{13}
\providecommand{\natexlab}[1]{#1}
\providecommand{\url}[1]{\texttt{#1}}
\expandafter\ifx\csname urlstyle\endcsname\relax
  \providecommand{\doi}[1]{doi: #1}\else
  \providecommand{\doi}{doi: \begingroup \urlstyle{rm}\Url}\fi

\bibitem[Basu \& Mitra(2020)Basu and Mitra]{basu2020deep}
Basu, Sanhita and Mitra, Sushmita.
\newblock Deep learning for screening covid-19 using chest x-ray images.
\newblock \emph{arXiv preprint arXiv:2004.10507}, 2020.

\bibitem[Cohen et~al.(2020)Cohen, Morrison, and Dao]{cohen2020covid}
Cohen, Joseph~Paul, Morrison, Paul, and Dao, Lan.
\newblock Covid-19 image data collection.
\newblock \emph{arXiv 2003.11597}, 2020.
\newblock URL \url{https://github.com/ieee8023/covid-chestxray-dataset}.

\bibitem[Jaiswal et~al.(2019)Jaiswal, Tiwari, Kumar, Gupta, Khanna, and
  Rodrigues]{jaiswal2019identifying}
Jaiswal, Amit~Kumar, Tiwari, Prayag, Kumar, Sachin, Gupta, Deepak, Khanna,
  Ashish, and Rodrigues, Joel~JPC.
\newblock Identifying pneumonia in chest x-rays: A deep learning approach.
\newblock \emph{Measurement}, 145:\penalty0 511--518, 2019.

\bibitem[Lakhani \& Sundaram(2017)Lakhani and Sundaram]{lakhani2017deep}
Lakhani, Paras and Sundaram, Baskaran.
\newblock Deep learning at chest radiography: automated classification of
  pulmonary tuberculosis by using convolutional neural networks.
\newblock \emph{Radiology}, 284\penalty0 (2):\penalty0 574--582, 2017.

\bibitem[Maguolo \& Nanni(2020)Maguolo and Nanni]{maguolo2020critic}
Maguolo, Gianluca and Nanni, Loris.
\newblock A critic evaluation of methods for covid-19 automatic detection from
  x-ray images.
\newblock \emph{arXiv preprint arXiv:2004.12823}, 2020.

\bibitem[Rajpurkar et~al.(2017)Rajpurkar, Irvin, Zhu, Yang, Mehta, Duan, Ding,
  Bagul, Langlotz, Shpanskaya, et~al.]{rajpurkar2017chexnet}
Rajpurkar, Pranav, Irvin, Jeremy, Zhu, Kaylie, Yang, Brandon, Mehta, Hershel,
  Duan, Tony, Ding, Daisy, Bagul, Aarti, Langlotz, Curtis, Shpanskaya, Katie,
  et~al.
\newblock Chexnet: Radiologist-level pneumonia detection on chest x-rays with
  deep learning.
\newblock \emph{arXiv preprint arXiv:1711.05225}, 2017.

\bibitem[RSNA(2018 (accessed July 2020))]{RSNA}
RSNA.
\newblock \emph{Kaggle RSNA Pneumonia Detection Challenge}, 2018 (accessed July
  2020).
\newblock URL
  \url{https://www.kaggle.com/c/rsna-pneumonia-detection-challenge}.

\bibitem[SIIM-ACR(2018 (accessed July 2020))]{SIIM-ACR}
SIIM-ACR.
\newblock \emph{SIIM-ACR Pneumothorax Segmentation}, 2018 (accessed July 2020).
\newblock URL
  \url{https://www.kaggle.com/c/siim-acr-pneumothorax-segmentation}.

\bibitem[Sirazitdinov et~al.(2019)Sirazitdinov, Kholiavchenko, Mustafaev,
  Yixuan, Kuleev, and Ibragimov]{sirazitdinov2019deep}
Sirazitdinov, Ilyas, Kholiavchenko, Maksym, Mustafaev, Tamerlan, Yixuan, Yuan,
  Kuleev, Ramil, and Ibragimov, Bulat.
\newblock Deep neural network ensemble for pneumonia localization from a
  large-scale chest x-ray database.
\newblock \emph{Computers \& Electrical Engineering}, 78:\penalty0 388--399,
  2019.

\bibitem[Tartaglione et~al.(2020)Tartaglione, Barbano, Berzovini, Calandri, and
  Grangetto]{tartaglione2020unveiling}
Tartaglione, Enzo, Barbano, Carlo~Alberto, Berzovini, Claudio, Calandri, Marco,
  and Grangetto, Marco.
\newblock Unveiling covid-19 from chest x-ray with deep learning: a hurdles
  race with small data.
\newblock \emph{arXiv preprint arXiv:2004.05405}, 2020.

\bibitem[Teichmann et~al.(2018)Teichmann, Weber, Zoellner, Cipolla, and
  Urtasun]{teichmann2018multinet}
Teichmann, Marvin, Weber, Michael, Zoellner, Marius, Cipolla, Roberto, and
  Urtasun, Raquel.
\newblock Multinet: Real-time joint semantic reasoning for autonomous driving.
\newblock In \emph{2018 IEEE Intelligent Vehicles Symposium (IV)}, pp.\
  1013--1020. IEEE, 2018.

\bibitem[Wang \& Wong(2020)Wang and Wong]{wang2020covid}
Wang, Linda and Wong, Alexander.
\newblock Covid-net: A tailored deep convolutional neural network design for
  detection of covid-19 cases from chest x-ray images.
\newblock \emph{arXiv preprint arXiv:2003.09871}, 2020.

\bibitem[Wang et~al.(2017)Wang, Peng, Lu, Lu, Bagheri, and
  Summers]{wang2017hospital}
Wang, X, Peng, Y, Lu, L, Lu, Z, Bagheri, M, and Summers, RM.
\newblock Hospital-scale chest x-ray database and benchmarks on
  weakly-supervised classification and localization of common thorax diseases.
\newblock In \emph{IEEE CVPR}, 2017.

\end{thebibliography}
\bibliographystyle{icml2017}

\end{document}